\documentclass{article}
\usepackage[a4paper,top=3.3cm,bottom=4.2cm,left=2.6cm,right=2.6cm]{geometry}

\usepackage{epsfig}
\usepackage{subcaption}
\usepackage{calc}
\usepackage{amssymb}
\usepackage{amstext}
\usepackage{amsmath}
\usepackage{amsthm}
\usepackage{multicol}
\usepackage{pslatex}
\usepackage{apalike}
\usepackage{algorithm2e}
\usepackage{algorithmic}

\usepackage[T1]{fontenc}
\usepackage{todonotes}
\usepackage{graphicx}
\usepackage{amsmath}
\usepackage{xcolor}
\usepackage[hidelinks]{hyperref}
\usepackage{url}
\usepackage{breakurl}
\usepackage{tikz}
\usetikzlibrary{shapes,arrows, calc}
\usepackage{svg}
\usepackage{float}
\usepackage{caption}
\usepackage{apacite}
\usepackage[bottom]{footmisc}
\usepackage{SCITEPRESS}     
\usepackage{blindtext}

\tikzstyle{decision} = [diamond, draw, fill=blue!20, 
    text width=4.5em, text badly centered, node distance=3cm, inner sep=0pt]
\tikzstyle{block} = [rectangle, draw, fill=blue!20, 
    text centered, rounded corners, minimum height=4em]
\tikzstyle{line} = [draw, ->]
\tikzstyle{invisible4} = [rectangle]
\tikzstyle{bidirectional} = [draw, <->]
\tikzstyle{arrow} = [draw, -latex',rounded corners]
\tikzstyle{cloud} = [draw, ellipse,fill=red!20, node distance=3cm,
    minimum height=2em]

\begin{document}

\urlstyle{same}

\title{Multidimensional Knowledge Graph Embeddings for International Trade Flow Analysis}

\author{
\authorname{%
Durgesh Nandini\orcidAuthor{0000-0002-9416-8554},%
Simon Blöthner\orcidAuthor{0009-0006-3462-4809},%
Mirco Schoenfeld\orcidAuthor{0000-0002-2843-3137} %
and Mario Larch\orcidAuthor{0000-0001-9355-2004}%
}
\affiliation{University of Bayreuth, Bayreuth, Germany}
\email{durgesh.nandini@uni-bayreuth.de}%
}



\abstract{Understanding the complex dynamics of high-dimensional, contingent, and strongly nonlinear economic data, often shaped by multiplicative processes, poses significant challenges for traditional regression methods
as such methods offer limited capacity to capture the structural changes they feature. To address this, we propose leveraging the potential of knowledge graph embeddings for economic trade data, in particular, to predict international trade relationships. We implement KonecoKG, a knowledge graph representation of economic trade data with multidimensional relationships using SDM-RDFizer and transform the relationships into a knowledge graph embedding using AmpliGraph. 
}

\onecolumn \maketitle \normalsize \setcounter{footnote}{0} \vfill

\makeatletter
\newcommand{\blindnote}[1]{%
  \def\@thefnmark{}
  \@footnotetext{#1}
}
\makeatother
\blindnote{All authors contributed equally.}

\section{Introduction}
\label{intro}

Knowledge graphs (KG) are repositories for factual information in triple form and have been increasingly prevalent across various domains. Exploring knowledge graph embedding models has emerged as a novel approach for exploiting knowledge graphs. These graphs have been useful, promoting numerous downstream tasks \cite{kun2023weext,abu2021domain}. These embeddings represent nodes and, in some cases, edges as continuous vectors, providing several advantages over traditional graph structures \cite{cai2018comprehensive, goyal2018graph, wang2017knowledge}. Beyond this, graph-based methodologies offer a promising avenue for capturing and quantifying narratives, particularly through knowledge graphs (KGs) which map interactions between concepts or events relevant to the research subjects \cite{wang2017knowledge, chen2020knowledge}. Numerous applications of these methods have demonstrated the efficacy of graph modelling and quantitative graph analysis in capturing complex economic relationships \cite{xia2021graph, chen2020review}.

Therefore, this study applies KG translational embedding techniques \cite{bordes2013translating} to solve inherent problems in empirical economic research. Economic research typically transforms the network of economic interactions into a format usable for (often even linear) inferential statistics or theoretical algebraic reasoning. However, this transformation can cause strong information and complexity compression, limiting the representativeness since the interaction and the underlying network structure have been almost completely ignored \cite{wolfram2002new}. Additionally, economic data has suffered from the problems of high-dimensionality, contingency and strong non-linearity, which originate from multiplicative dynamics \cite{donoho2000high, bolon2016feature, raudenbush2002hierarchical}. This paper discusses these issues when further analysing economic data in Section \ref{challenges}.

To address these problems, we propose that every economic interaction can be represented within a network structure. In the latter, we establish the concept of an economic trade network as a system of interconnecting countries based on their trade relations. Our primary aim is to explore the predictive capabilities inherent within such a network, specifically focusing on forecasting flows between country pairs.
To do this, we introduce KonecoKG, a downstream KG embedding model featuring multidimensional translational relationships for the international economic bilateral data. A multidimensional relationship in the context of KGs is one between entities encompassing multiple attributes or interactions simultaneously. Such a relationship offers various advantages because it facilitates capturing the combined effect of multiple attributes rather than one single entity-attribute relation. For example, a simple binary relationship might indicate only a single type of link, e.g., \textit{``country A trades with country B''}, On the other hand, a multidimensional relationship captures a richer set of associations, such as fixed effects, and contextual information like trade volumes, geographical proximity and economic indicators. The latter include gross domestic product (GDP) and population size. By incorporating these diverse dimensions into the relationships, a KG can provide a more nuanced, comprehensive representation of the data, enabling more accurate and insightful analysis using the embedding model. By leveraging a trade network dataset, we anticipate future trade opportunities by integrating historical trade patterns with insights into the trading behaviours of neighbouring countries within the network. Additionally, multi-attributes such as trade agreements, geographic proximity and economic similarities act as network features to refine the accuracy of our predictions. The main contributions of this work are as follows:
\begin{itemize}
    \item Establish a trade network as a graph representation of countries with relationships indicating trade flows, eliminating the problems of non-linearity and non-hierarchical representations in international economic bilateral trade data.
    \item Introduce the KonecoTradeFlow ontology, which represents the concepts of the international economic bilateral trade data.
    \item Introduce KonecoKG, a downstream graph embedding model that applies translational techniques to forecast trade flows.
\end{itemize}

To the best of our knowledge, our study is one of the few pioneering efforts in utilising a large-scale economic trade network to predict trade flows between countries. The implications of accurately forecasting trade dynamics are significant, offering valuable insights for policymakers, businesses and investors to optimise international trade strategies. Additionally, the study identifies emerging market trends and encourages economic growth \cite{anand2021innovation}. 

The rest of the paper is organised as follows: First, Section \ref{lit} gives an overview of the literature on conventional econometric approaches and graph-based methods and underlines the key challenges in economic research, establishing the significance of the current study’s contribution. Having outlined the challenges, Section \ref{approach} focuses on the approach we are using and describes our process to construct the TradeFlow ontology, the embedding methods used, and the learning strategy. Section \ref{experiments} focuses on the experimental setup and the evaluation metrics used. Section \ref{results} provides insights into the results obtained and discusses their implications. Lastly, we highlight the findings of the research and conclude in Section \ref{conclusion}, whereby we also layout ideas for future research in this field.

\section{State of the Art}
\label{lit}

In this section, we discuss the challenges associated with the economic data comprising the foundation of this research. Additionally, this section reviews current methods used to address these challenges and identifies gaps in these methods, including those involving KGs, to underscore the necessity of the proposed approach.

\subsection{Challenges of Economic Data}
\label{challenges}
Many formal, data-driven efforts do not adequately address the unique characteristics of economic data \cite{schumpeter1933common}. Economic exchanges are shaped by subjectivity \cite{menger1871principles}, creating context dependence and contingency, sometimes called localised knowledge \cite{hayek1945use}. Together, these characteristics hinder people from gathering reliable insights from economic data. Multi- or high-dimensionality requires incorporating many variables into models, which must be capable of untangling all the non-linear interactions between these variables. Beyond this, many economic variables of interest exhibit strong power law behaviour, also called heavy- or fat-tailed behaviour  \cite{gabaix2009power, di2011power, axtell2001zipf, hinloopen2006comparative}. This process produces a slow convergence speed, leaving one in a world of pre-asymptotics with estimates which have not yet reached stable, reliable values. Even if such a value is reached, it is often unrepresentative of individual observations due to the large difference in magnitude  \cite{taleb2020statistical}. 

\begin{figure}[!htbp]
    \centering
    \includegraphics[width = 0.46\textwidth]{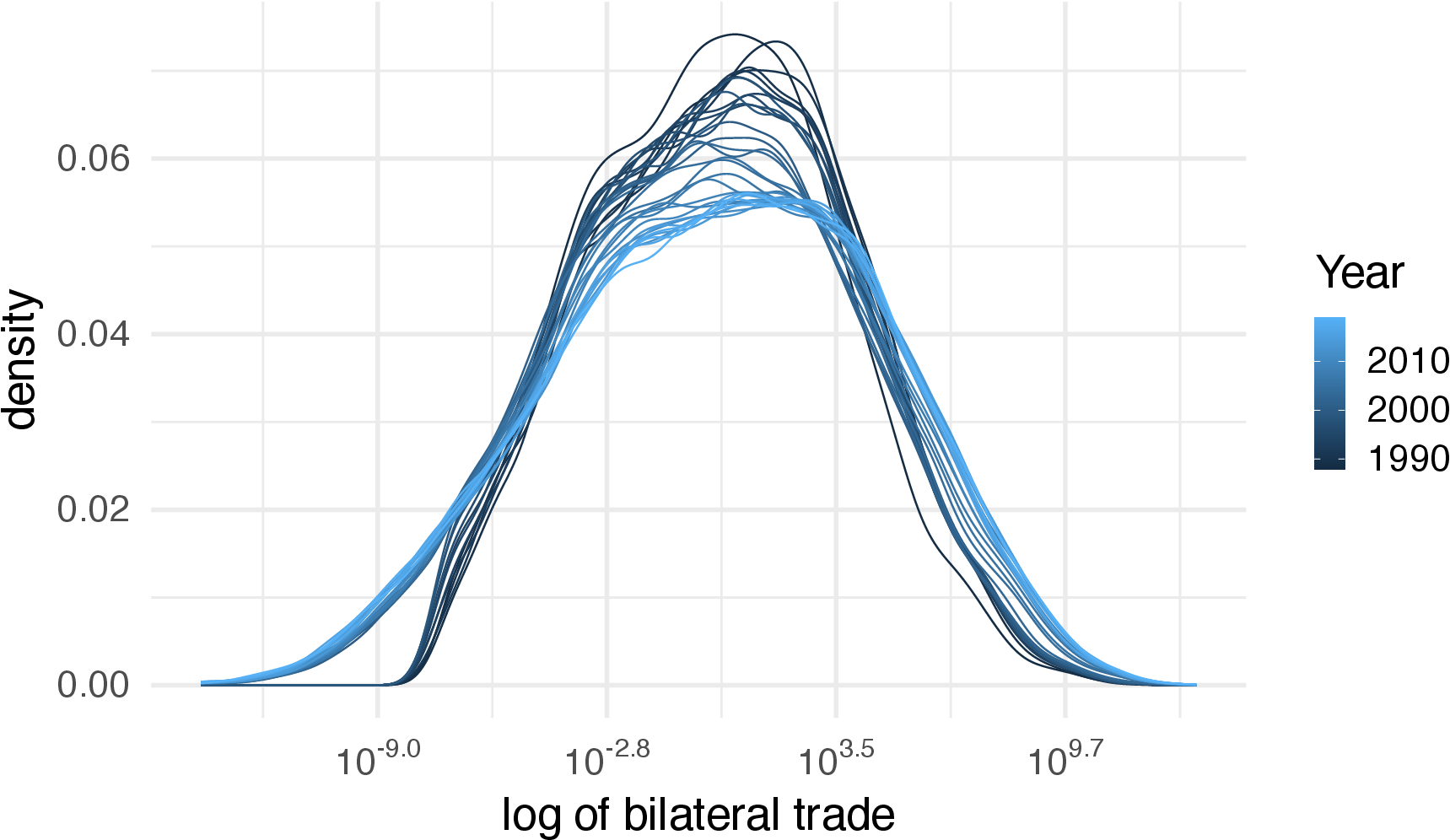}
    \caption{Log density of bilateral trade flows across time}
    \label{fig:log_dists}
\end{figure}

Figure \ref{fig:log_dists} exemplifies this characteristic. Looking at all the bilateral trade flows grouped by year shows that the data has much heavier tails than a Gaussian distribution, also called the normal distribution. This can be seen by the mass of probability in the tails, as opposed to that in the body of the distribution. Notably, the distributions in Figure \ref{fig:log_dists} are on a logarithmic scale, making the problem exponentially more pronounced. The distribution has a tail index $\alpha \approx 1$, leading to slow convergence and imprecise estimates.

All these phenomena are expressed to the highest degree when dealing with international trade flows, as they necessarily aggregate all the individual choices to the highest possible level \cite{blothner2022economic}. 

\subsection{Methods for Economic Data Analysis} The standard empirical approach in economic data analysis, a field referred to as econometrics, is a regression model. To explain variations in bilateral international trade flows, the workhorse model is to estimate the theory-founded gravity equation using the Poisson pseudo-maximum likelihood (PPML) estimator \cite{silva2006log,head2014gravity,yotov2016advanced}. Generally, these approaches rely on a large set of fixed effects to control for unobservable effects in various dimensions. This process includes dummy variables for every country, and sometimes for every country pair, as well as exporter-year and importer-year observations \cite{fally2015structural, egger2016glm}. We will also rely on this specification when comparing it to our KG model in section \ref{results}.
Another approach is the descriptive analysis of networks such as in \cite{de2011world, basile2018impact}. However, such work does not facilitate inference or the understanding of factors that drive certain characteristics within the network. Recent advances in informatics, especially the combination of machine learning models with graph structures, can provide new insights into the field of economics. However, due to their focus on causal explanation, traditional economic analysis methods have predominantly relied on linear models and supervised learning techniques.

\subsection{Knowledge Graph for Economic Trade Flow Data} Relevant recent advances have been made in the field of neural networks and KG networks. \cite{sellami2024harnessing} used a Graph Convolution Network for predicting the trade relation between countries. Elsewhere, \cite{rincon2023enhancing} used a synthetic triple-generation algorithm for enhancing downstream tasks in KG embeddings based on the graph complement. \cite{rincon2023accurate} leveraged KG embeddings for modelling international trade, focusing on link prediction using embeddings, and explored the integration of traditional machine learning methods with KG embeddings. \cite{meng2022retracted} used an enterprise KG to predict China’s Free Trade Zone. \cite{gastinger2023dynamic} used a KG to explain trade patterns among various countries. Other approaches to this process have been in the economic trade flow data analysis including economic planning \cite{shao2017bidirectional}, and industrial economic status \cite{quan2022visualization}.

\section{Methodology}
\label{approach}
\begin{figure}[!htbp] 
\includegraphics[width=0.47\textwidth]{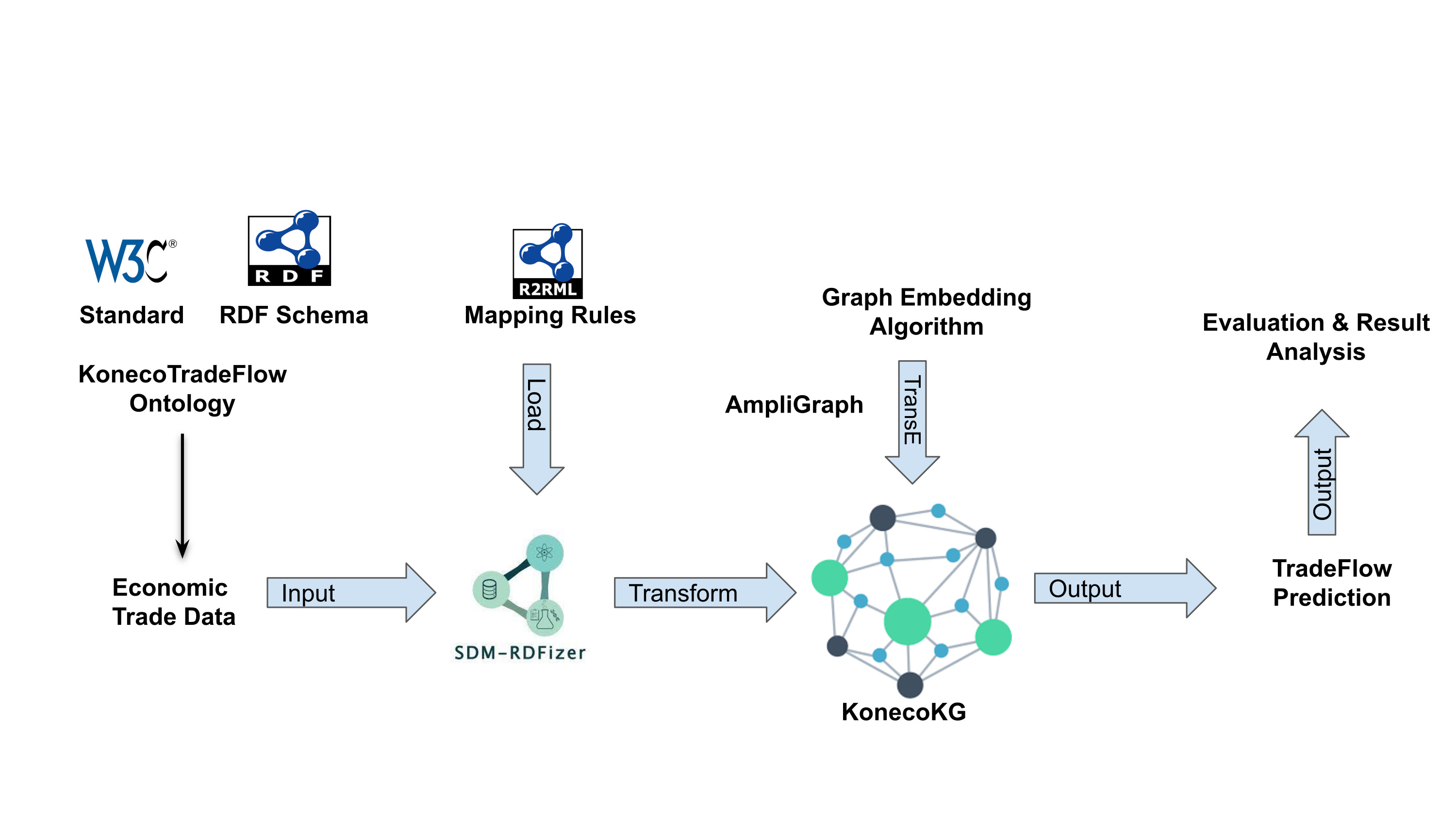}
\caption{Trade flow prediction and analysis pipeline} \label{fig:methdology}
\end{figure}

This section explains the creation of KonecoKG, applying embedding techniques, and predicting trade values. KonecoKG takes triples in the form of Subject ($\mathsf{s}$), Predicate ($\mathsf{p}$) and Object ($\mathsf{o}$) as inputs for multiple relationships, and then forms embedding vectors for each relation. Next, the embedding vectors are combined as an average embedding vector to predict trade flows between countries as the final output. Figure \ref{fig:methdology} shows a diagrammatic representation of the methodology. The subsections here provide an extensive overview of the methodology followed. 

\subsection{International Economic Trade Flow Data} The initial step entails identifying relevant aspects of the dataset. 
Using trade data from \cite{borchert2021international}, spanning 1986 to 2016, and encompassing 170 countries, we tackle the questions of economic drivers of trade flows. To determine this, we added explanatory data from \cite{gurevich2018dynamic} for GDP and population data, and \cite{mayer2011notes} for information on geographic distances between countries. Lastly, we employed data about trade agreements from \cite{EggerLarch2008}, a strong predictor of international trade flows. We aggregated this data into a tabular format, leaving us with over 2.5 million observations over the whole time frame.

\subsection{Data Processing and Feature Selection} 

A detailed explanation of selected features is given in Table \ref{table:datapropoerties}, comprising the key determinants of international trade. Economic theory predicts that larger countries, measured using population or economic size (GDP), are more able than smaller countries to trade with each other. Specifically, country size affects a country’s division of labour and thus the ‘roundaboutness’ of production or how many intermediary capital goods for production are employed. As this number grows, countries develop greater potential to trade. In contrast, countries facing high trade costs will trade less. In contrast, countries facing high trade costs trade less. These costs can be either direct because they are far apart (distance, geographic position) or indirect due to other trade barriers which increase the transaction cost (triangulation, trust, transfer). 

\subsection{Data Modelling as KonecoTradeFlow Ontology} 

\begin{figure}[!htbp] 
\resizebox{\linewidth}{!}{
    \begin{tikzpicture}
    \centering
        \node [block, fill=green!20] (init)  at (0, 0) {Country};
        \node (TradeRelation) at (0,1.5){trade relation};
        \node [block, fill=yellow!20, minimum height=0.6cm, minimum width=2.9cm] (distance)  at (0, 2.7) {distance};
        \node [block, fill=yellow!20, minimum height=0.6cm, minimum width=2.9cm] (trade)  at (-3.5, 2.7) {trade};
        \node [block, fill=yellow!20, minimum height=0.6cm, minimum width=2.9cm] (tradeAgreement)  at (3.5, 2.7) {trade agreement};
        \node [block, fill=yellow!20, minimum height=0.6cm, minimum width=2.9cm] (GDPwdi)  at (3.5, -1.2) {GDP (WDI)};
        \node [block, fill=yellow!20, minimum height=0.6cm, minimum width=2.9cm] (GDPpwt)  at (3.5, -2.0) {GDP (PWT)};
        \node [block, fill=yellow!20, minimum height=0.6cm, minimum width=2.9cm] (population)  at (3.5, -2.8) {population};
        \node [block, fill=yellow!20, minimum height=0.6cm, minimum width=2.9cm] (latitude)  at (3.5, -3.6) {latitude};
        \node [block, fill=yellow!20, minimum height=0.6cm, minimum width=2.9cm] (longitude)  at (3.5, -4.4) {longitude};
        \node at (1,-1.0) {\scriptsize has GDP (WDI)};
        \node at (1,-1.8) {\scriptsize has GDP (PWT)};
        \node at (1,-2.6) {\scriptsize has population};
        \node at (1,-3.4) {\scriptsize has latitude};
        \node at (1,-4.2) {\scriptsize has longitude};

        \path[line] (init.east) -- ++(0.5,0) -- ++(0,1) -- ++(-2.5,0) -- ++(0,-1) |-  (init.west);
        \path[line] (TradeRelation.east) -- ++(2.29,0) -|(tradeAgreement.south);
        \path[line] (TradeRelation.west) -- ++(-2.29,0) -| (trade.south);
        \path[line] (TradeRelation.north) -| (distance.south);
        \path [line] (init.south) -- ++(0,-1) |-  (GDPwdi.west);
        \path [line] (init.south) -- ++(0,-1.8) |-  (GDPpwt.west);
        \path [line] (init.south) -- ++(0,-2.6) |-  (population.west);
        \path [line] (init.south) -- ++(0,-3.5) |-  (latitude.west);
        \path [line] (init.south) -- ++(0,-3.5) |-  (longitude.west);
      
    \end{tikzpicture}
    }
    \caption{KonecoKG data model diagram}
    \label{fig:classdiagram}
\end{figure}
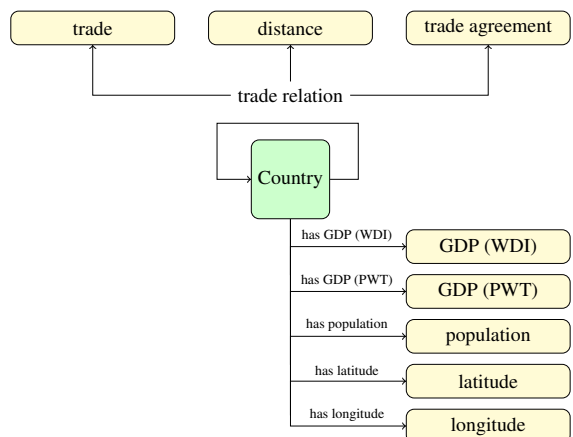

The subsequent step in the construction of the model involved creating a formal semantic representation of the dataset to serve as a structured framework for organising and categorising concepts, entities and relationships. The advantage to this method is that it captures the hierarchical structure and dependencies among these features, allowing for a nuanced understanding of their interplay in shaping trade dynamics \cite{chandrasekaran1999ontologies,fensel2001ontologies,uschold1996ontologies}. Figure \ref{fig:classdiagram} represents the hierarchical structure of our data as a class diagram. From the figure, we identify \textit{`trade relation'} as our main class. A complete list of data properties \cite{uschold1996ontologies,chandrasekaran1999ontologies} and object properties \cite{uschold1996ontologies,chandrasekaran1999ontologies} is provided in Table \ref{table:datapropoerties}.

\begin{table*}
\centering
\caption{Data \& object properties and their description}\label{table:datapropoerties}
\begin{tabular}{|l|l|}
\hline
\textbf{Data Property} &  \textbf{Description} \\
\hline

trade &  volume of bilateral trade  \\
distance & geodesic distance between the exporter and importer \\
trade agreement & whether a trade agreement exists between two countries \\
GDP (WDI) & GDP of a country as measured by the World Development Indicators\\
GDP (PWT) & GDP of a country as measured by the Penn World Tables\\
population & population of a country\\
latitude & geographical latitude of a country\\
longitude & geographical longitude of a country\\
\hline
\textbf{Object Property} &  \textbf{Description} \\
\hline

tradesWith & indicates whether a trade relation exists between two countries  \\
\hline
\end{tabular}
\end{table*}

\subsection{Knowledge Graph Construction} In the next step, we use the KonecoTradeFlow ontology formulated in the above step to a structured representation in a KG, producing a set of triples. 

To this end, we converted our dataset into KonecoKG using 
 
SDM-RDFizer, an open-source tool and interpreter of the W3C Recommendations Standard R2RML\footnote{https://www.w3.org/TR/r2rml/} and its RDF Mapping Language (RML)\footnote{https://rml.io/specs/rml/} extension used for the semantification process 
and used in KG creation in prior research \cite{shahi2023fakekg}. The RDF is a standardised data model used to describe resources on the web using subject-predicate-object statements, known as triples. Each triple comprises three components: subject, predicate, and object. The following are the detailed steps used to convert data into KonecoKG: 
       
\begin{itemize}
\item \textbf{Entity identification:} we identified the entities or resources that we wanted to represent in RDF. For our use case, the entities were countries, specifically the exporters and the importers, their associated data properties, and their relationships among them. 

\item \textbf{Ontology:} We used the KonecoTradeFlow ontology as vocabulary to model trade data. For instance, the data property \textit{trade} represents trade (in millions of US Dollars).

\item \textbf{Mapping Rule:} Following the steps of SDM-RDFizer, mapping rules were created using R2RML. We assigned the base URL as \textit{www.koneco.de}, and mapping rules assigned the data values to the corresponding subjects, predicates, and objects in the RDF triples and assign the appropriate Uniform Resource Identifier (URI) for the entities and properties and linked them to represent the relationships. For instance, the trade column of the dataset is mapped as \textit{tradeValue}. A snapshot of the data properties from the KonecoTradeFlow ontology is given below. \\

\begin{algorithmic}
\STATE \textit{rr:predicateObjectMap [}
\STATE \textit{\hspace{1em} rr:predicate kg:tradeValue;}
\STATE \textit{\hspace{1em} rr:objectMap [}
\STATE \textit{\hspace{2em} rml:reference "trade"}
\STATE \textit{\hspace{1em} ]}
\STATE ]
\end{algorithmic}

\vspace{1em}
\item \textbf{Serialising as RDF:} We serialised the RDF triples into a specific RDF serialisation format. We used the Turtle format to store and exchange RDF data while preserving the structure and semantics of the triples.

\end{itemize}

We represented \textit{Facts} in a KG as relationships between entities — for instance, \textit{<ARB\_NZL hasTradeValue n>}, means Aruba exports, goods and services of value \textit{n} to New Zealand. We build a series of such statements derived from the raw data collection to represent them as a graph.

\subsection{Knowledge Graph Embeddings} 

After KonecoKG is created, we employed KG embeddings,  generating embedding scores for each triple, thus encoding entities and relationships into numerical vectors. In this way, the model processes intricate patterns and semantic information as a continuous vector space, facilitating enhanced effective analysis and inference. Next, we trained the triples, derived from the KG, using three embedding models. Specifically, we employed TransE \cite{bordes2013translating}, ComplEx \cite{trouillon2016complex}, and DistMult \cite{dettmers2018convolutional}. 
\par{The \textit{TransE} is a deterministic approach which regards the relation as a translation operation from the head entity to the tail entity and utilises a distance-based scoring function to measure the plausibility of triples. Each of the latter offers unique advantages and facilitates different perspectives on capturing the semantics of the underlying data. On the other hand, the \textit{ComplEx} and \textit{DistMult} utilise tensor factorisation and model the interaction of entities and relations by vector-matrix product to obtain the expressive power of the data.}

\subsection{Prediction Model} 
This section explains this study's approach to finding trade relations using link prediction in KonecoKG. Link prediction is the process of exploiting the existing facts in a KG to infer missing ones. For triples <$\mathsf{s}$,$\mathsf{p}$,$\mathsf{o}$> in KonecoKG, where <$\mathsf{s}$> refers to a country pair, <$\mathsf{p}$> represents countries' trade relation, and <$\mathsf{o}$> represents the monetary value of the trade occurring between two countries. Then we used tail prediction to predict the values of o.

Subsequently, we adopted a corruption-based learning strategy \cite{bordes2013translating} to make predictions. This strategy entailed intentionally introducing corruptions or perturbations to the input data during the training process to enhance the model's ability to generalise and make accurate predictions. The rationale behind this approach is its ability to encourage the model to learn robust representations of the data resilient to noise and variations. Exposing the model to a diverse range of corrupted inputs during training caused it to become more adept at discerning meaningful patterns and relationships from the data, thus improving its predictive performance on unseen or noisy data. 
 
Practically, the corruption strategy can be implemented by augmenting the training dataset with artificially corrupted samples or by introducing random perturbations to the input data during each training iteration. The degree and type of corruption introduced can be tailored based on the specific characteristics of the dataset and the desired robustness of the model. We have expanded on the use of the corruption model, adopted by us, in Section \ref{experiments}.

\subsection{Evaluation}
\label{eval}
We evaluated the quality of the embedding model by measuring how well the model could complete facts. The prediction model predicted the tail of all the possible facts of KonecoKG.

We evaluated the embedding model using the Mean Reciprocal Rank (MRR) and Hits@N. Once the best embedding model was determined, we applied the Mean Squared Error (MSE) metric to calculate the error in the predicted values. 

\begin{itemize}
    \item \textbf{MRR} measures how well the model ranks the correct entity or relation among the candidates in the predicted list by measuring the average of the reciprocal ranks of the correct tail entities across all test triples. If the correct tail entity is ranked first, the reciprocal rank is 1; if it is ranked second, the reciprocal rank is 1/2, and so on. MRR is defined as:

          \[
    \text{MRR} = \frac{1}{|\text{Test Triples}|} \sum_{i=1}^{|\text{Test Triples}|} \frac{1}{\text{Rank}_i}
    \]

    \item \textbf{Hits@N} measures the proportion of test triples where the correct answer appears within the top N predictions. Similar to MRR, we have a set of test triples and a ranked list of candidate tail entities for each test triple. This metric calculates the percentage of test triples for which the correct tail entity appears within the top N ranks in the predicted list. Hits@N is defined as follows:

    \[
    \text{HITS@N} = \frac{\text{Number of Hits at Rank } \leq N}{|\text{Test Triples}|}
    \]
        
    \item \textbf{MSE} is used to measure the error in the prediction model by computing the average squared difference between estimated trade values ($\hat{y}_i$) and actual trade values ($y_i$). MSE is defined as follows:
    \[
    \text{MSE} = \frac{1}{n} \sum_{i=1}^{n} (y_i - \hat{y}_i)^2
    \]

    \end{itemize}

\section{Experimental Setup}
\label{experiments}

To begin with, we utilised Protégé\footnote{https://protege.stanford.edu/}, a widely used ontology editor 
and followed ontology design approach \cite{dutta2015mod,dutta2015yamo}, 
to build and visualise the KonecoTradeFlow ontology. The importance of this initial step before any other experimental setup was to provide insights into formalising the concept for mapping the trade flow data to create KonecoKG. This initial step was crucial to provide a visual representation of the relationships between different entities, helping to clarify how various types are connected and ensuring consistent data structure. They also served to define the formal relationships between concepts and offered a shared understanding of the domain enabling reasoning over the data.

In the second step, to simplify the start of the experimental process, we first filtered out data for each year from the entire dataset collection since the data comprises of trade information over a span of time. We did not deal with the temporal aspect of the data, rather we created a separate Kg for each year.

The third step was the conversion of trade flow data into a format suitable for KG embedding. To perform this, we employed the SDM-RDFizer. We started by formulating the required R2RML mapping rules in Turtle\footnote{https://www.w3.org/TR/turtle/}. In the rules, we specified the classes, properties, and relationships we aim to necessitate in the graph. We used the mapping rules to generate <$\mathsf{s}$,$\mathsf{p}$,$\mathsf{o}$> triples, also in the Turtle format. The result of all the triples (subdivisions of Classes, and Relationships) is the KonecoKG, which is also in the Turtle format. 

\begin{figure}
\centering
\includegraphics[width=0.45\textwidth]{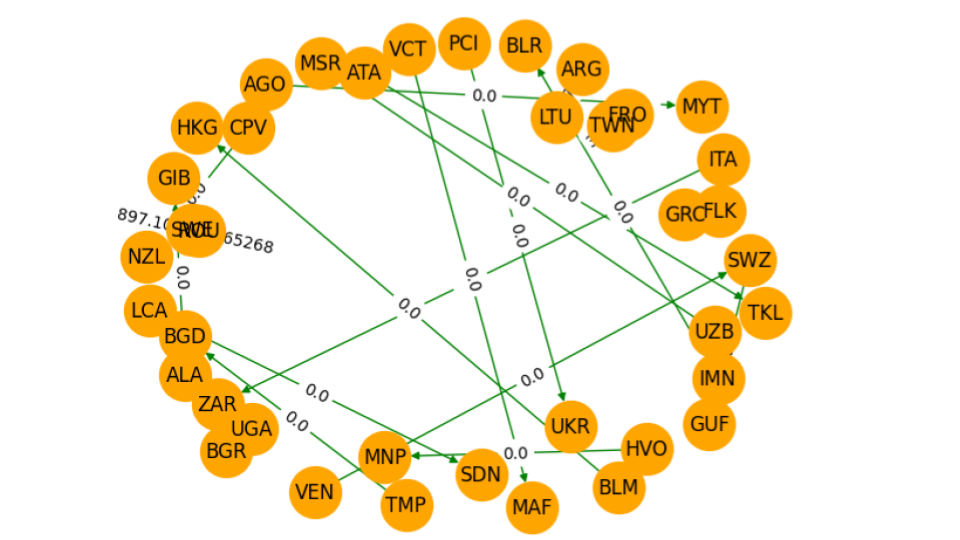}
\caption{Sample trade network in KonecoKG} \label{fig:tradenetwork}
\end{figure}

In the fourth step, we used the generated Turtle output to parse the graph using the RDFLib\footnote{\url{https://rdflib.readthedocs.io/en/stable/apidocs/rdflib.html}} graph package in our model. Figure \ref{fig:tradenetwork} shows a simplified glimpse of the trade network. In the figure, the nodes represent the countries and the edges represent a bilateral trade relationship between two countries. The labels of the edges represent the value of the monetary trade exchange in millions of US Dollar. A value of 0.0 indicates that there is no trade relation at all. The origin of the edge represents the export country, and the direction represents the import country. 

In the fifth step, we employed the AmpliGraph \cite{costabello2019ampligraph} Python library\footnote{https://github.com/Accenture/AmpliGraph} to process the graph and to transform it into a vectorised multidimensional representation of the statements it contained. Several potential embedding model architectures were available through the AmpliGraph package with a variety of parameters. As mentioned in Section \ref{approach}, to evaluate the performance of KG embedding models, we experimented with three algorithms: TransE, CompleX, and DistMult. To optimise the model parameters, we employed a grid search methodology, systematically exploring various combinations to identify the most effective settings. Table \ref{table:modeleval} presents the metric performance scores of the models obtained rounded off to the third decimal place. 

Notably, our experiments revealed that the TransE model consistently outperformed the alternatives by 10\%, for our data. Although ComplEx outperformed the other models for Hits@1 and Hits@10, however, upon further experiments, we found that when the N in Hits@N increased, the model's performance consistently decreased. On the other hand, with an increase in N, the Hits@N score for TransE consistently increased. Therefore, we decided to go ahead with the TransE model for further experimentation. We used the TransE to predict the trade values and evaluate the performance metrics. 


In general, the model trains by comparing statements ($\mathsf{s}$,$\mathsf{p}$,$\mathsf{o}$) known to be true against statements likely to be false based on local closed-world assumptions. This strategy measured the distances between different statements and aimed to minimise the said distance. An essential component of this experimental strategy is the corruption algorithm. The corruption algorithm creates negative triples by corrupting a true triple either by replacing the head or the tail entity with a random incorrect entity. This forces the model to distinguish between true and false facts, thereby enhancing model robustness.

Initially, we utilised the default corruption method provided by TransE. However, recognising the potential benefits of introducing controlled noise into the training process, we subsequently modified this strategy by corrupting trade values by a relative value of 20\% of their true values. Through experimentation (20\%, 50\%, 70\%, 100\%, 120\%), we determined that a corruption level of about 20\% optimally enhanced the results. This adjustment appeared to simulate real-world variations and uncertainties in trade dynamics, thereby improving the model's ability to generalise to unseen data.

Subsequently, we trained our model for 1000 epochs, with an embedding size of 150 dimensions. These settings were chosen based on preliminary experiments and empirical observations to strike a balance between model performance and computational efficiency, ensuring timely convergence and effective learning. Tables \ref{subtable:outparameters} and \ref{subtable:inparameters} provides a full overview of the parameter values. 

However, in our analysis, we noticed that we had to change the hyperparameters for a comparable prediction for the in-sample and out-of-sample methods, most notably in the epoch and batch size. The in-sample method required fewer epochs and lower negative sampling for predicting trade flows. Table \ref{subtable:inparameters} provides a full overview of the parameter values for in-sampling.

The trained model works by generalising relationships not yet seen by the neural network to predict the likelihood of a relationship being true with a given confidence.

\begin{table}[!htbp]
\begin{minipage}[t]{0.46\textwidth}
\centering
\caption{Out-of-sample embedding parameters}\label{subtable:outparameters}
\begin{tabular}{|l|l|}
\hline
Parameter &  Value\\
\hline
Epochs &  1500\\
Embedding size &  150\\
Corruptions & 30\\
Batch size & 30\\
Loss function & Pairwise\\
Initialiser & Xavier\\
Regulariser & LP, 'lambda': 0.01, 'p': 2\\
Optimiser & Adam\\ 
Learning rate & 0.001\\
\hline
\end{tabular}
\end{minipage}
\hspace{0.48\textwidth}
\vspace{3mm}
\begin{minipage}[t]{0.46\textwidth}
\centering
\vspace{2mm}
\caption{In-sample embedding parameters}\label{subtable:inparameters}
\begin{tabular}{|l|l|}
\hline
Parameter &  Value\\
\hline
Epochs &  1000\\
Embedding size &  150\\
Corruptions & 10\\
Batch size & 50\\
Loss function & Pairwise\\
Initialiser & Xavier\\
Regulariser & LP, 'lambda': 0.01,'p': 2\\
Optimiser & Adam\\ 
Learning rate & 0.001\\
\hline
\end{tabular}
\end{minipage}
\end{table}

\begin{table*}[!htbp]
  \centering
  \caption{Results of trade flow prediction}\label{table:modeleval}
  \begin{subtable}{.3\linewidth}
    \centering
    \caption{ComplEx}\label{table:complexeval}
    \begin{tabular}{|l|l|}
      \hline
      Metric &  Score\\
      \hline
      MRR &  0.483\\
      Hits@1 &  \textbf{0.428}\\
      Hits@10 & \textbf{0.513}\\
      Hits@100 & 0.512\\ 
      Hits@1000 & 0.592\\
      \hline
    \end{tabular}
  \end{subtable}%
  \begin{subtable}{.3\linewidth}
    \centering
    \caption{TransE}\label{table:transeeval}
    \begin{tabular}{|l|l|}
      \hline
      Metric &  Score\\
      \hline
      MRR &  \textbf{0.587}\\
      Hits@1 &  0.298\\
      Hits@10 & 0.459\\
      Hits@100 & \textbf{0.576}\\ 
      Hits@1000 & \textbf{0.719}\\
      \hline
    \end{tabular}
  \end{subtable}%
  \begin{subtable}{.3\linewidth}
    \centering
    \caption{DistMult}\label{table:distmulteval}
    \begin{tabular}{|l|l|}
      \hline
      Metric &  Score\\
      \hline
      MRR &  0.376\\
      Hits@1 &  0.311\\
      Hits@10 & 0.404\\
      Hits@100 & 0.491\\
      Hits@1000 & 0.504\\
      \hline
    \end{tabular}
  \end{subtable}
\end{table*}

\section{Results and Discussion}
\label{results}

To evaluate the effectiveness of our model on unseen data, we applied the leave-one-out cross-validation. We iterated over each country relation as the test set and used the rest of the chunk as the training set. Thus, we reported the average scores of the runs, each consisting of 1000 epochs. Then, we used the performance metrics MRR \cite{costabello2019ampligraph}, and Hits@N \cite{costabello2019ampligraph} to evaluate the predictions generated by the model. As described in Section \ref{approach} we experimented with hyperparameters of three KG embedding models, namely, ComplEx, TransE and DistMult. 

Furthermore, we also compared our results with a baseline regression model 
using the Mean Squared Error(MSE) metrics. We enlist the Mean Squared Error (MSE) (in millions) comparison in Table \ref{tab:MSE}.

\begin{table}[!htbp]
\centering
\caption{Mean Squared Error by model}\label{tab:MSE}
\begin{tabular}{|l|l|}
\hline
Model &  Mean Squared Error (in million)\\
\hline
PPML  &  2256.65 \\ 
ComplEx  &  256.65 \\ 
DistMult  &  196.26 \\ 
\textbf{TransE} & \textbf{14.493564}\\
\hline
\end{tabular}
\end{table}

\begin{figure}[!htbp]
    \centering
    \begin{subfigure}{0.45\textwidth}
        \centering
        \includegraphics[width=\linewidth]{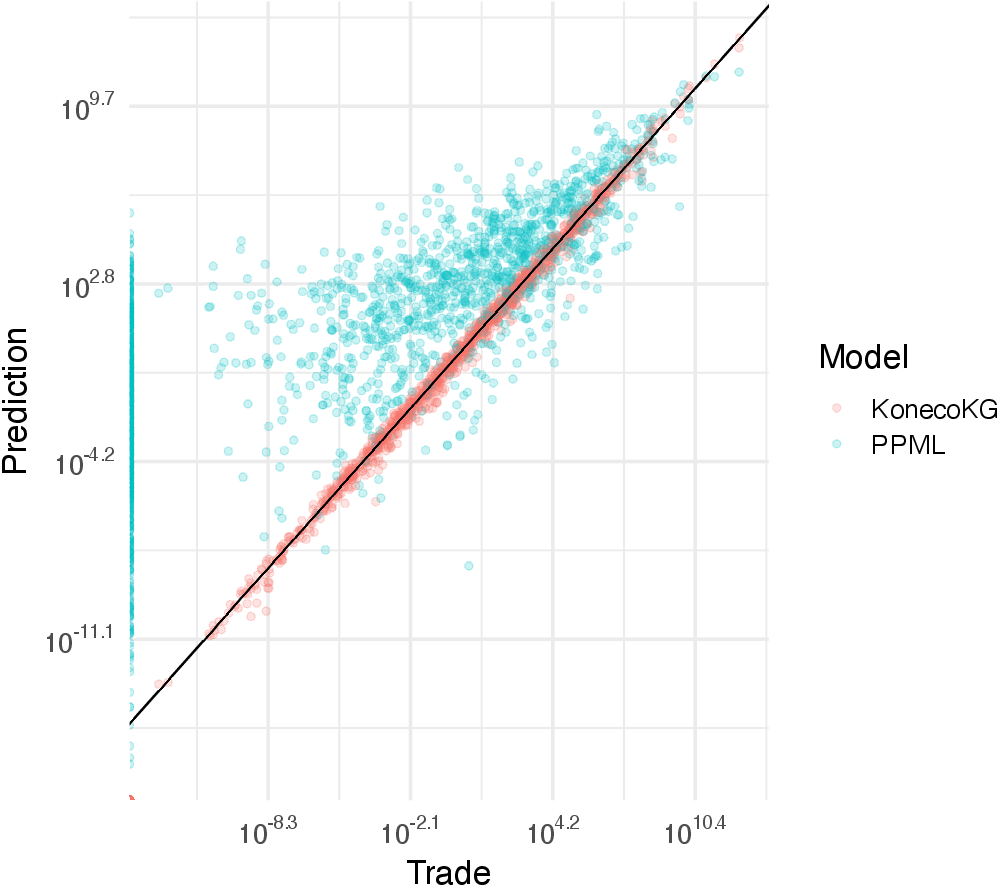}
        \caption{Out-of-sample}
        \label{fig:out_sample}
        \vspace{2.5mm}
    \end{subfigure}
    \hfill
    \begin{subfigure}{0.45\textwidth}
        \centering
        \includegraphics[width=\linewidth]{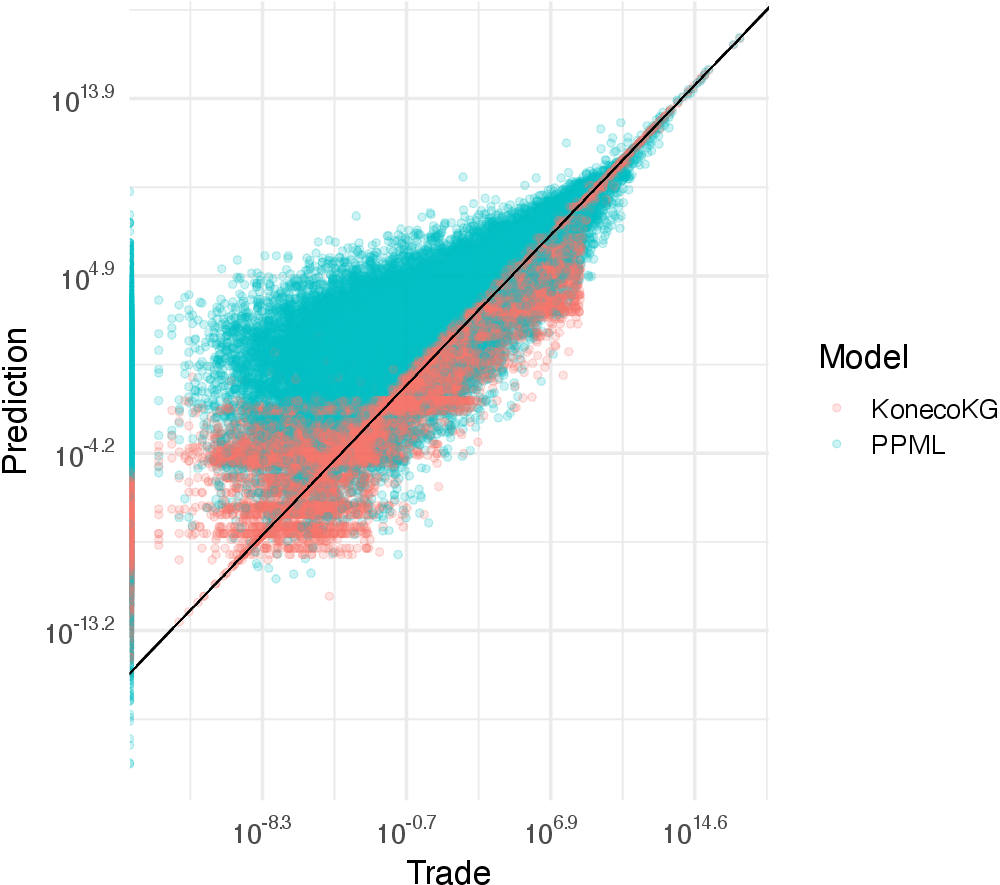}
        \caption{In-sample}
        \label{fig:in_sample}
        \vspace{2.5mm}
    \end{subfigure}
    \caption{Comparison of predictions (on log-log scale)}
    \label{fig:predictions}
\end{figure}

Lastly, we applied the traditional approach, for instance, PPML for predicting the trade value along with the proposed approach. This model vastly outperformed the conventional models in out-of-sample prediction tasks. Relying on the MSE, it is up to $155$ times better than a comparable estimate using PPML with fixed effects, as seen from Table \ref{tab:MSE}. In this vein, Figure \ref{fig:out_sample} shows that the machine learning approach predicts values at every scale quite accurately. Notably the 45$^{\circ}$ line represents a perfect fit. Even in the in-sample case, KonecoKG outperforms PPML by a factor of 50. PPML is biased towards large values, which is a commonly observed result. Furthermore, our model predicts all the $0$ trade flows correctly, a feature which is impossible for PPML. 

Generally, conventional regression-based approaches aim to ascertain the average response of a variable of interest to a change, usually in policy. In our case, this process could involve signing a trade agreement between one or multiple countries. As motivated in Section \ref{challenges} and reinforced by our results, the influence of certain factors is mediated by a plethora of contingent factors. Even if such an average response could be achieved, a complex interplay of dependencies could make the individual experience of an economic agent, such as a country, to differ wildly from the estimated average. This phenomenon has recently been addressed in economics \cite{peters2019ergodicity}. For this reason, we see great potential for graph-based learning algorithms to untangle the complexities at the heart of economic processes and to deepen our understanding of economic relationships.


\section{Conclusion and Future Work}
\label{conclusion}

In this work, we applied KG embedding techniques to predict trade flows in the international bilateral trade flow data by formulating KonecoKG, a downstream model. A significant advantage of introducing graph structure is that it alleviates the problems of non-linearity and hierarchical high-dimensional data. The proposed approach outperforms the state-of-the-art model in predicting trade values from 50, for in-sample tasks, to 155 times, for out-of-sample tasks. Currently, this approach has been applied for a limited number of properties. 

Additionally, this approach can be extended by combining KGs built from other data sources which are nearly impossible to include in standard approaches, due to their unstructured nature. These sources could include text-based agreements, news, exchange and auction-based data, and market phenomena such as decentralised finance. 

An alternate and immediate subsequent extension of the work would be explaining the embedding and the prediction model to identify the key determinants of the model. Additionally, post-hoc explainability models could be used to explore the results obtained. 

Another possible step could be the use of the time dimension. As time is the medium through which any economic process is realised, this would offer a much more realistic picture. That is, much of recent econometric research has used this feature, facilitating the path dependence of multiplicative processes.

\section*{Supplemental Material}

The raw data can be viewed and downloaded from {\textit{Mario Larch's} Regional Trade Agreements Database}\footnote{https://www.ewf.uni-bayreuth.de/en/research/RTA-data/index.html}, {Dynamic Gravity Dataset}\footnote{https://www.usitc.gov/data/gravity/dgd.htm}, {International Trade and Production Database for Estimation (ITPD-E)}\footnote{https://www.usitc.gov/data/gravity/itpde.htm}. In particular, we will release the ontology model, mapping rules for creating the KonecoTradeFlow ontology, code to tune hyperparameters for the ComplEx, TransE, and DistMult, code to train, and predict model using TransE. The project, data, and the Python Code can also be found at the GitHub\footnote{https://github.com/durgeshnandini/Multidimensional-Knowledge-Graph-Embeddings-for-International-Trade-Flow-Analysis}.

\section*{Acknowledgement}
The work has been done as the part of KONECO project, and it has received funding from the Bundesministerium f{\"u}r Bildung und Forschung (BMBF) under grant No 16DKWN095.

We also thank Rebekka Koch, our student assistant, for her valuable efforts in collecting related literature, and for efforts during various crucial parts of the paper development.

\bibliographystyle{apacite}
{\small
\bibliography{example}}

\end{document}